\documentclass[conference]{IEEEtran}
\usepackage[T1]{fontenc}
\IEEEoverridecommandlockouts
\usepackage{bm}
\usepackage{cite}
\usepackage{amsmath,amssymb,amsfonts}
\usepackage{algorithmic}
\usepackage{textcomp}
\usepackage{graphicx}
\usepackage{tikz}
\usepackage{pgfplots}
\usepackage{caption}
\usepackage{subcaption}
\usepackage{stfloats}

\def\BibTeX{{\rm B\kern-.05em{\sc i\kern-.025em b}\kern-.08em
    T\kern-.1667em\lower.7ex\hbox{E}\kern-.125emX}}
\begin{document}

\title{Decentralized Fusion of 3D Extended Object Tracking based on a B-Spline Shape Model}

\author{
\IEEEauthorblockN{
    Longfei Han\IEEEauthorrefmark{1} \IEEEauthorrefmark{2}, 
    Klaus Kefferp\"utz\IEEEauthorrefmark{1} \IEEEauthorrefmark{3}, 
    and J\"urgen Beyerer\IEEEauthorrefmark{2} \IEEEauthorrefmark{4}
}
\IEEEauthorblockA{\IEEEauthorrefmark{1} Application Center \guillemotright Connected Mobility and Infrastructure\guillemotleft , Fraunhofer IVI, Ingolstadt, Germany}
\IEEEauthorblockA{\IEEEauthorrefmark{2} Vision and Fusion Laboratory, Karlsruhe Institute of Technology (KIT), Karlsruhe, Germany}
\IEEEauthorblockA{\IEEEauthorrefmark{3} Technische Hochschule Ingolstadt, Ingolstadt, Germany}
\IEEEauthorblockA{\IEEEauthorrefmark{4} Fraunhofer IOSB, Karlsruhe, Germany}
}

\maketitle

\begin{abstract}
Extended Object Tracking (EOT) exploits the high resolution of modern sensors for detailed environmental perception. 
Combined with decentralized fusion, it contributes to a more scalable and robust perception system.
This paper investigates the decentralized fusion of 3D EOT using a B-spline curve based model.
The spline curve is used to represent the side-view profile, which is then extruded with a width to form a 3D shape. 
We use covariance intersection (CI) for the decentralized fusion and discuss the challenge of applying it to EOT.
We further evaluate the tracking result of the decentralized fusion with simulated and real datasets of traffic scenarios.
We show that the CI-based fusion can significantly improve the tracking performance for sensors with unfavorable perspective.
\end{abstract}


\section{Introduction}
Extended Object Tracking (EOT) allows autonomous agents to better estimate the state of dynamic objects in their environment by allowing sensors to generate more than one measurement from a single object and track it with a shape model \cite{eot}. 
Most of the proposed EOT methods track an object as a contour in a 2D plane with different geometric models, ranging from ellipses (e.g. \cite{RandomMatrix}), or superellipses (e.g. \cite{deformablesuper}) through star-convex models (e.g. \cite{RHMStar, GP}), to even flexible models like B-spline \cite{bspline}. 
To cope with perception sensors that provide 3D environmental measurements, 3D shape models have also been proposed \cite{3dextrusion, 3dgp, NURBS, FCDS}.
These models mainly gain the extra dimensional description at the expense of the increased number of parameters. 
A new method is proposed in \cite{sideview} where a B-spline curve is used to describe the side view profile of the object in detail.
The width of the object is then represented by a single parameter to keep the number of parameters small. 
This method is well suited for tracking objects such as vehicles in traffic scenarios.

Many of the works are performed with a single sensor or in a centralized fusion scheme, assuming the raw sensor data are first shared and the tracker performs the state estimation with them.
Although the fusion of data, for example, from the roof of an autonomous vehicle and from a roadside infrastructure pole of an intelligent transportation system complements the viewpoints and improves the tracking, centralized fusion could suffer from the failure of a single node and the high transmission cost of data traffic.
In contrast, decentralized fusion allows agents to perform state estimation with their local sensor data and then share their results, including uncertainties, with each other.
This approach can provide a flexible, scalable, and robust system solution because the agents are independent of each other, the system can be easily scaled, and single point failure can be avoided.
\begin{figure}[t]
\centering
        \begin{tikzpicture}
        \node[anchor=south west,inner sep=0] (image) at (0,0)             {\includegraphics[width=0.48\linewidth]{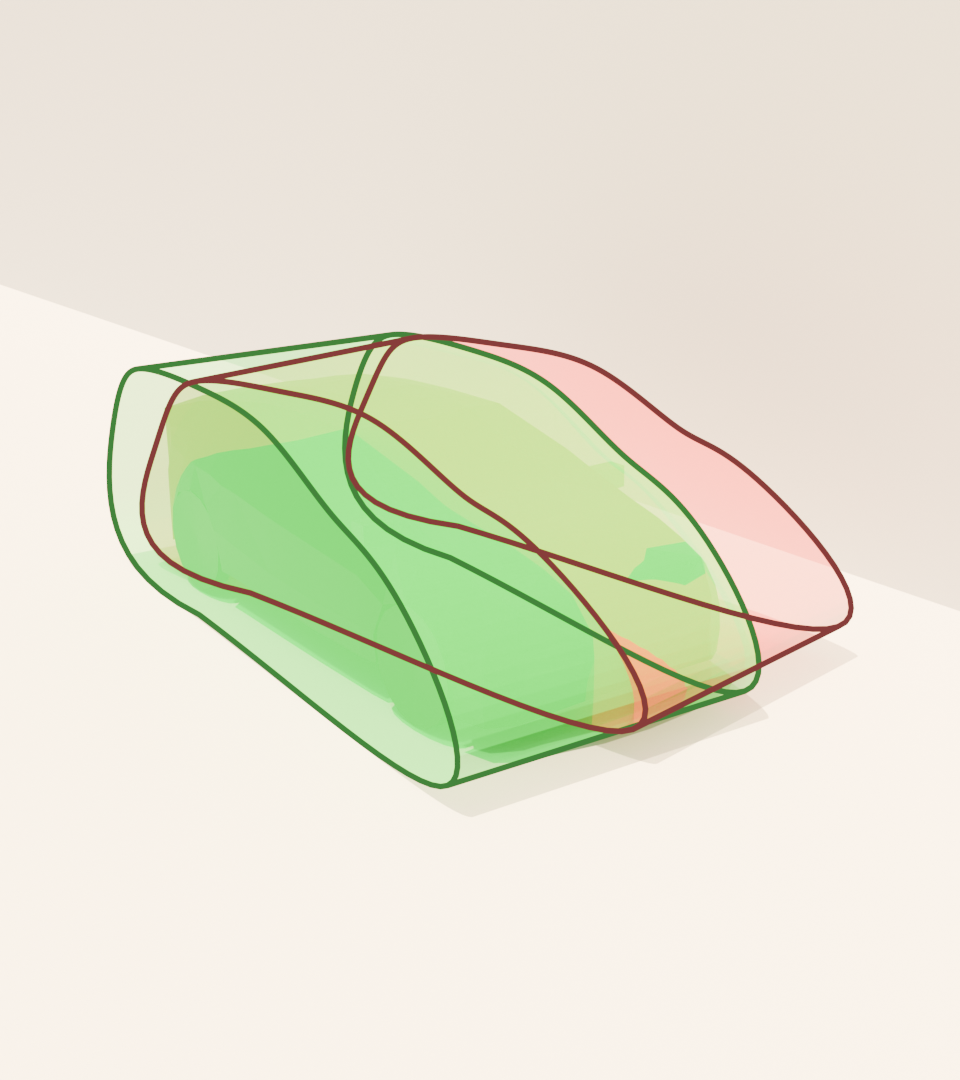}
        \includegraphics[width=0.48\linewidth]{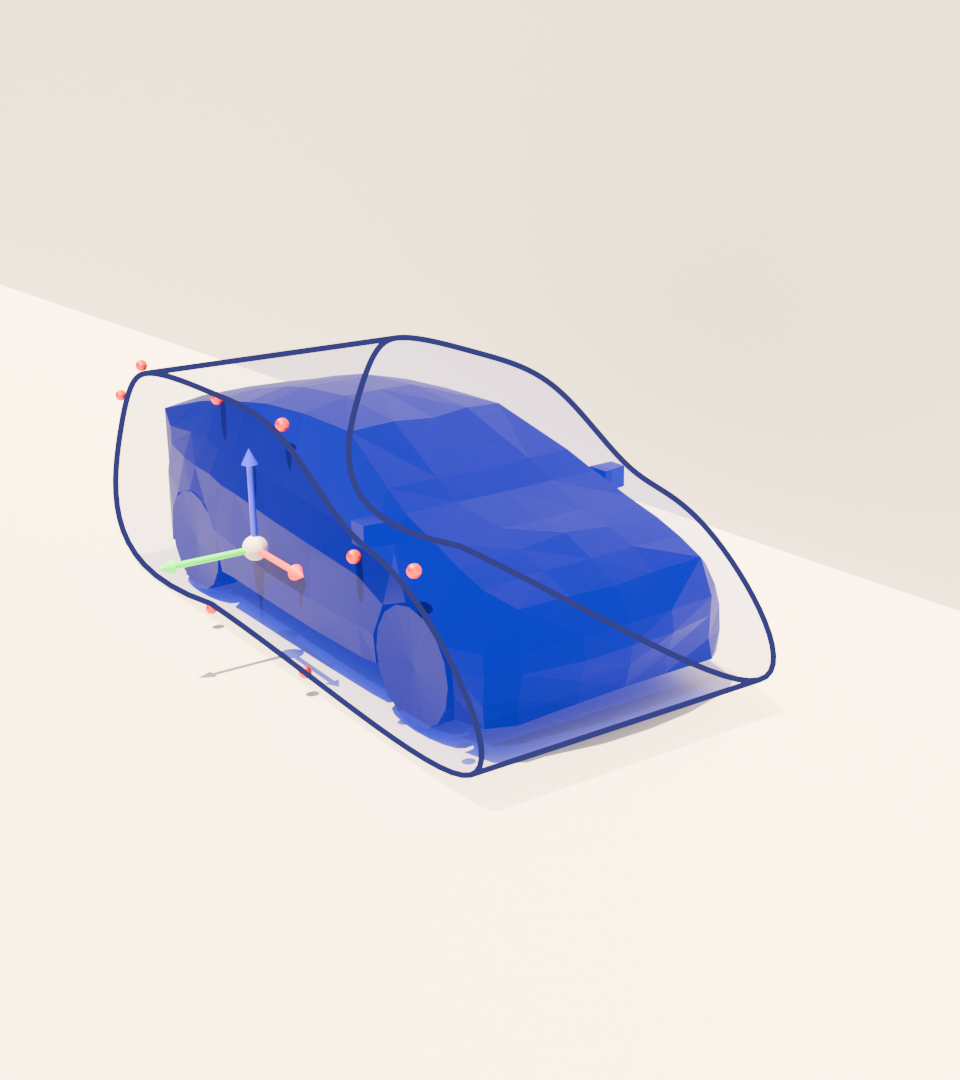}};
        \begin{scope}[x={(image.south east)},y={(image.north west)}]
            \draw[green, thick, ->] (0.15, 0.2) node[below=-2pt] {Estimation 1} -- (0.2, 0.35); 
            \draw[red, thick, ->] (0.38, 0.8) node[above=-2pt] {Estimation 2} -- (0.35, 0.55); 
            \draw[->, ultra thick] (0.4,0.5)  -- (0.6,0.5) node[midway, above]{Fusion};
            \draw[blue, thick, ->] (0.8,0.8) node[above=-2pt] {Fused Estimation} -- (0.72,0.67);    
            \draw[black, thick, ->] (0.7,0.2) node[below=-2pt] {Groud Truth State} -- (0.8,0.40); 
        \end{scope}
    \end{tikzpicture} 
    \caption{
    The concept of decentralized fusion of 3D EOT using the model of extruding a side-view profile represented by a B-spline curve.
    The estimation 1 and estimation 2 are computed locally.
    The fusion is performed using CI.
    The ground truth state of the object is shown together with its mesh.
    The points in the right image represent the control points of the B-spline curve.
    }
    \label{fig: concept}
\end{figure}
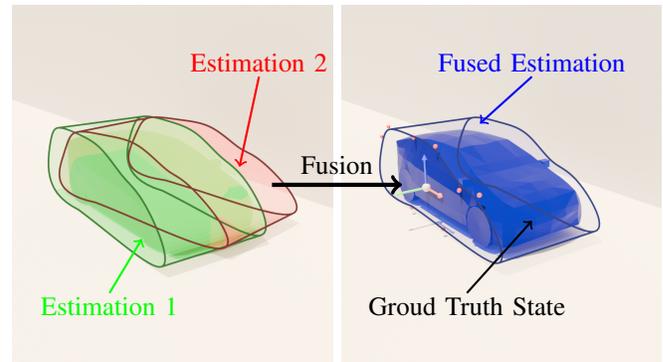

For decentralized tracking, one of the most profound methods is Covariance Intersection (CI) \cite{CIJulier, CIrevisited, CItutorial}.
When the correlation between the estimates is unknown, CI fusion yields an optimal result.
However, this result is also conservative. 
If the correlation is further analyzed, the result can be improved, which gives the need for derivatives like split CI \cite{splitCI} or inverse CI \cite{ICI}.
Different shape models should be an important consideration when applying decentralized fusion in EOT.
The authors in \cite{decentralizedRM, decentralizeDiffusion} use the diffusion strategy in the fusion of the ellipse extended objects.
Random Ellipse Density is introduced in \cite{decentralizedRED} for track-to-track fusion of ellipses. 
In an effort to extend the decentralized EOT work to more complex shape models, Fröhle et al. propose to perform the fusion of the posterior estimation of a Gaussian Process (GP) based EOT tracking using CI \cite{decentralizedGPPMBM}. 
To the best of the author's knowledge, there hasn't been any work on decentralized fusion of 3D EOT.

In this work, we investigate the fusion of decentralized 3D EOT using the newly proposed tracking model \cite{sideview}. 
We analyze the performance of this tracking method in fusion considering the challenges of decentralized EOT tracking. 
We then integrate CI into the tracking workflow for the fusion of posterior estimates from two trackers (as shown in Fig. \ref{fig: concept}).
Finally, we evaluate the decentralized fusion on both simulated data and a real-world dataset, showing the improvement of the tracking result compared to a single tracker.

The rest of the paper is organized as follows: Section \ref{chapter: problemformulation} describes the problem formulation by introducing the B-spline model, the tracker, and the CI.
Section \ref{chapter: implementation} discusses the CI approach within the proposed tracking framework and the implementation. 
Section \ref{chapter: evaluation} gives an evaluation of the decentralized fusion on both simulated data and real open datasets.
Section \ref{chapter: conclusion} concludes the paper and gives further research directions.
\section{Problem Formulation}
\label{chapter: problemformulation}
This section illustrates the problem of decentralized fusion of 3D extended object tracking based on a B-spline curve model for side view profile representation.
First, we give a brief introduction to B-spline and B-spline curves. 
Then we describe the state estimation by extruding the side view profile represented by a B-spline curve.
Finally, covariance intersection is briefly introduced.
\subsection{B-spline curve}
Let $\bm{\underline{c}}=(\bm{c}_i)_{i=1}^n$\footnote{Bold symbols denote vectors, underlined symbols denote sets of points.} represent a set of control points and $\bm{\tau}=(\tau_i)_{i=1}^{n+d+1}$ a knot vector, we can define a B-spline curve $\bm{s}(\tau)$ of degree $d$ as:
\begin{equation}
\label{eq: spline}
    \bm{s}(\tau) = \sum_{i = 1}^{n} B_{i,d}(\tau)\bm{c}_{i},
\end{equation}
where $\tau$ is a "walking parameter" along the spline curve.
The function $B_{i,d}(\tau)$ is further defined by recursive expression as 
\begin{equation}
    B_{i,d}(\tau) = \frac{\tau-\tau_i}{\tau_{i+d}-\tau_{i}}B_{i,d-1}(\tau) + \frac{\tau_{i+1+d}-\tau}{\tau_{i+1+d}-\tau_{i+1}}B_{i+1,d-1}(\tau).\end{equation}
In cases of  ``divided by zero'', we let the function $B_{i,d}$ return zero.
The basic function $B_{i,d}(\tau) = B_{i,d, \bm{\tau}}(\tau)$ is called a B-spline of degree $d$ (with knots $\bm{\tau}$).
Given this definition, a B-spline curve can be used to represent the side view profile of a vehicle, depending on the chosen control points. 
We refer the reader to \cite{splinemethod} for B-spline and its properties.

\subsection{3D state estimation using B-spline curves}
\label{chapter: shapemodel}
Following \cite{sideview}, an object is modeled as an extrusion of the side view profile represented by a B-spline curve. 
It has a state representation at time step $k$ as
\begin{equation}
\label{eq: state}
   \bm{x}_k= [\bm{x}_{k,m}^\top, \bm{x}_{k,e}^\top]^\top 
\end{equation}
By neglecting the time index $k$ and extending $\bm{x}_m$ and $\bm{x}_e$, we have 
\begin{equation}
    \bm{x}_m = [x_x, x_y, v_{xy}, \psi, \omega, x_z, v_z]^\top,
\end{equation}
\begin{equation}
 \bm{x}_e = [q, x_{c^1_x}, x_{c^1_z}, \cdots, x_{c^n_x}, x_{c^n_z} ]^\top.
\end{equation}
The position of the object in 3D is $[x_x, x_y, x_z]^\top$.
It moves with a velocity of $v_{xy}$ in the 2D plane and with a velocity of $v_z$ in the vertical direction.
The orientation of the object together with its angular velocity is represented by $\psi$ and $\omega$. 
For the shape representation, the parameter $q$ is the width of the object for extrusion. 
The set of control points of the 2D B-spline curve in the $xz$ plane $\underline{\bm{x}}_c =(\bm{x}_{c^i})_{i=1}^n=([x_{c^i_x}, x_{c^i_z}]^\top)_{i=1}^n $ is used to control the side view profile.

For tracking, an extended Kalman filter can be derived given the nonlinearity in the model.
The process model \begin{equation}
    \bm{x}_{k+1} =\bm{f}(\bm{x}_{k} ) + \bm{w}
\end{equation} consists of 3 parts and their corresponding process noise $\bm{w}$:
The motion in the $xy$ plane is modeled with a constant turn rate and velocity (CTRV) \cite{comparison} process.
The motion in the vertical direction $z$ is modeled with a constant velocity (CV) process.
Finally, the shape parameters are modeled with a very low noise random walk.

The object extent is represented by the extrusion of the 2D B-spline curve and two caps an the ends of the extrusion. 
All possible points $\bm{z}^L$ (measurement sources) on the surface in the body frame of reference are:
\begin{equation}
\label{eq: shapemodel}
    \bm{z}^L(\bm{x}_e, \tau) =  \begin{cases}
        \left[z^L_x(\tau),z^L_y, z^L_z(\tau) \right]^\top,  \quad z^L_y \in \left[-\frac{q}{2}, \frac{q}{2}\right],\\
        \left[z_x^L, z_y^L, z_z^L\right]^\top, \quad z_y^L \in \left\{ \pm \frac{q}{2} \right\} .
    \end{cases}
\end{equation}
The points on the extrusion are represented with the B-spline:
\begin{equation}
\left[z^L_x(\tau),  z^L_z(\tau)\right]^\top= \bm{s}(\tau) = \sum_{i = 1}^{n} B_{i,d}(\tau)\bm{x}_{c^i}. 
\end{equation}
For the points on the caps, we have \begin{equation}
\left[z^L_x,z^L_z\right] \in \Gamma = \left\{ \left[z^L_x,z^L_z\right] \ | \ g(z^L_x,z^L_z) \leq 0 \right\}    
\end{equation}
as the set of interior points on the cap, $g$ denotes a signed distance function. 

In a measurement model we have $\bm{y}_{k,l}$ generated from $\bm{z}_{k,l}$ at time step $k$ indexed by $l$:
\begin{equation}
\label{eq: y}
    \bm{y}_{k,l} = \bm{z}_{k,l} + \bm{v}_{k,l}, \quad \bm{v}_{k,l} \sim \mathcal{N}(\bm{0}, \bm{R}_{k,l}),
\end{equation}

The shape model can further be represented in a level-set \cite{levelset}$:     \mathcal{L}_\phi(\bm{x_k},t) = \{ \bm{z}_{k,l} | \phi(\bm{x}_k, \bm{z}_{k,l})=t\} $ with $t=0$, indicating that the measurements are generated from the surface of the object.
We can then write the pseudo-measurement model as
\begin{equation}
    \begin{split}
    0&=t - \phi(\bm{x}_k, \bm{y}_{k,l} - \bm{v}_{k,l}) \\
    &=0 - \phi(\bm{x}_k, \bm{y}_{k,l} - \bm{v}_{k,l}) \\
    &= h(\bm{x}_k, \bm{y}_{k,l}, \bm{v}_{k,l}).
    \end{split}
\end{equation}
In the body frame of reference, we use the $x,z$ component of the points on the extrusion, and the $y$ components of the points in the caps. 
The pseudo-measurement 0 then has the representation as follows:
\begin{equation}
\label{eq:h}
    \begin{split}
    \bm{0}&=\bm{h}=[h_{e,x} \ h_{e,z}, \ h_{c,y}]^\top,\\
        h_{e,x}&=\left(T^{-1}[\bm{y}_{k,l}^\top \ ,1]^\top\right)_x - \bm{s}_x(\tau_{k,l}) - \bm{v}_{k,l,x}\\
        &=\mathrm{c}(\psi_k) y_{k,l,x} + \mathrm{s}(\psi_k)y_{k,l,y} - \mathrm{c}(\psi_k)x_{k,x}\\
        &\quad - \mathrm{s}(\psi_k)x_{k,y} -     \sum_{i = 1}^{n} B_{i,d}(\tau_{k,l})x_{c_x^i} - \bm{v}_{k,l,x},\\        
        h_{e,z}&=\left(T^{-1}[\bm{y}_{k,l}^\top \ ,1]^\top\right)_z - \bm{s}_z(\tau_{k,l}) - \bm{v}_{k,l,z}\\
        &=y_{k,l,z} -x_{k,z}- \sum_{i = 1}^{n} B_{i,d}(\tau_{k,l})x_{c_z^i} - \bm{v}_{k,l,z},\\
        h_{c,y}&=\left(T^{-1}[\bm{y}_{k,l}^\top \ ,1]^\top\right)_y - \left(\pm \frac{q}{2}\right) - \bm{v}_{k,l,y}\\
        &=-\mathrm{s}(\psi_k) y_{k,l,x} + \mathrm{c}(\psi_k)y_{k,l,y} + \mathrm{s}(\psi_k)x_{k,x}\\
        & \quad- \mathrm{c}(\psi_k)x_{k,y}
         - \left(\pm \frac{q}{2}\right) - \bm{v}_{k,l,y},\\
    \end{split}
\end{equation}
The measurement model can be linearized around the estimation point by computing its Jacobi matrix. 
The corresponding measurements can then be used for the correction step (see \cite{sideview} for more details.)
\subsection{Covariance intersection}
Let $\hat{\bm{x}}^1$ und $\hat{\bm{x}}^2$ be two state estimations\footnote{We use superscript to represent different sensors or different sources.}, where
\begin{equation}
    \begin{split}
    \hat{\bm{x}}^1 &= \bm{x} + \tilde{\bm{x}}^1, \bm{P}^1 = \mathbf{E}[\tilde{\bm{x}}^1(\tilde{\bm{x}}^1)^{\mathrm{T}}] \\
    \hat{\bm{x}}^2 &= \bm{x} + \tilde{\bm{x}}^2, \bm{P}^2 = \mathbf{E}[\tilde{\bm{x}}^2(\tilde{\bm{x}}^2)^{\mathrm{T}}]
    \end{split}
\end{equation}
Here we have $\tilde{\bm{x}}^i = \hat{\bm{x}}^i-\bm{x}^i$ as error in the estimations. 
The matrix $\bm{P}^i$ is the error covariance matrix of the $i$th estimate.
The matrix $\bm{P}^{12} = \mathbf{E}[\tilde{\bm{x}}^1(\tilde{\bm{x}}^2)^{\mathrm{T}}]$ is the cross-covariance matrix, which is used to describe correlations between $\hat{\bm{x}}^1$ and $\hat{\bm{x}}^2$.
The fusion requires the calculation of $\bm{K}$ and $\bm{L} = \bm{I}- \bm{K}$ to get  
\begin{equation}
    \hat{\bm{x}}^{\mathrm{fus}} = \bm{K}\hat{\bm{x}}^1 + \bm{L}\hat{\bm{x}}^2.
\end{equation}
The optimal error covariance matrix with known $\bm{P}^{12}$ is
\begin{equation}
\label{eq: cross-covariance}
\begin{split}
    \bm{P}^{\mathrm{fus}} &= \mathbf{E}[\tilde{\bm{x}}^{\mathrm{fus}}(\tilde{\bm{x}}^{\mathrm{fus}})^{\mathrm{T}}]
    = \bm{K}\bm{P}^1\bm{K}^{\mathrm{T}} + \bm{K}\bm{P}^{12}\bm{L}^{\mathrm{T}} + \\ &\bm{L}(\bm{P}^{12})^{\mathrm{T}}\bm{K}^{\mathrm{T}} + \bm{L}\bm{P}^2\bm{L}^{\mathrm{T}}.
\end{split}
\end{equation}

Given the fact that the correlation can come from different sources, e.g. the estimates are affected by common process noise, or the same information is included in the estimate by the fusion in the previous time step, or the interferences happen between the sensors \cite{CItutorial}, we usually cannot get true correlations $\bm{P}^{12}$ to compute an optimal fusion.
CI can then be used as a conservative optimal solution:
\begin{equation}
\label{eq: ci}
\begin{split}
    \hat{\bm{x}}^{\mathrm{CI}} &= \bm{P}^{\mathrm{CI}}(\omega(\bm{P}^1)^{-1}\hat{\bm{x}}^1 + (1-\omega)(\bm{P}^2)^{-1}\hat{\bm{x}}^2),\\
    \bm{P}^{\mathrm{CI}} &= (\omega(\bm{P}^1)^{-1} + (1-\omega)(\bm{P}^2)^{-1})^{-1},
\end{split}
\end{equation}
where $\omega \in [0,1]$ is found by optimizing a cost function $C(\bm{P}^{\mathrm{CI}})$, usually defined as the trace or determinant.

\section{Implementation}
\label{chapter: implementation}
The fusion of extended object states is not trivial.
The introduction of orientation can lead to counterintuitive phenomena in the fusion. 
As illustrated in \cite{decentralizedRED}, if the orientation difference between two estimates is large (e.g., $\pi$), the result could become $\pi/2$, leading the estimate to incorrectly rotate $90^\circ$.
Orientation estimation combined with shape model parameterization could also cause problems. 
For example, the radial function with fixed angular resolution in \cite{GP} needs to be handled carefully, especially if the angular resolution matches the orientation difference between two estimates, the association of the basis points in the fusion could be disturbed when viewed from the global frame.
Furthermore, some shape parameters have unequal influences in the modeling, such as the Fourier coefficients in \cite{RHMStar}.
The fusion of the lower-order coefficients could dramatically change the shape, overriding the influence of the higher-order coefficients.

The 3D EOT method in section \ref{chapter: shapemodel} is designed taking into account the above mentioned challenges.
First, the orientation is incorporated into the 2D motion, which limits the orientation to around the evolving direction of the measurements.
This limits the difference between possible estimates to a reasonable value.
Second, the control points of the B-spline curve are defined flexibly in the body frame with no fixed resolution and the orientation does not directly affect their positions.
Their order is used for association in the fusion.
The fused result could again be a freely placed control point.
Third, each control point in the shape parameterization has only a local influence on the curve, which ensures a balanced contribution of the parameters after the fusion.

Our decentralized fusion of EOT is implemented with CI under the assumption that the correlation between the two systems is unknown.
The implementation is done by inserting the state (Eq. (\ref{eq: state})) together with its uncertainty into Eq. (\ref{eq: ci}). 
The cost function $C(\bm{P}^{\mathrm{CI}})$ is defined as $\det(\bm{P}^{\mathrm{CI}})$.
In this work, we study the theoretical result of the fusion and perform the implementation in a single process, neglecting the practical aspect such as the data transmission in a network.
The fusion is conducted when two sensors have performed tracking based on their measurement data.
If only one sensor has sufficient measurements, the fusion will not take place.

\section{Evaluation}
\label{chapter: evaluation}

\begin{figure}[t]
\centering
        \begin{tikzpicture}
        \clip (0,25pt) rectangle (\linewidth,110pt);
        \node[anchor=south west,inner sep=0] (image) at (0,0){\includegraphics[width=1\linewidth]{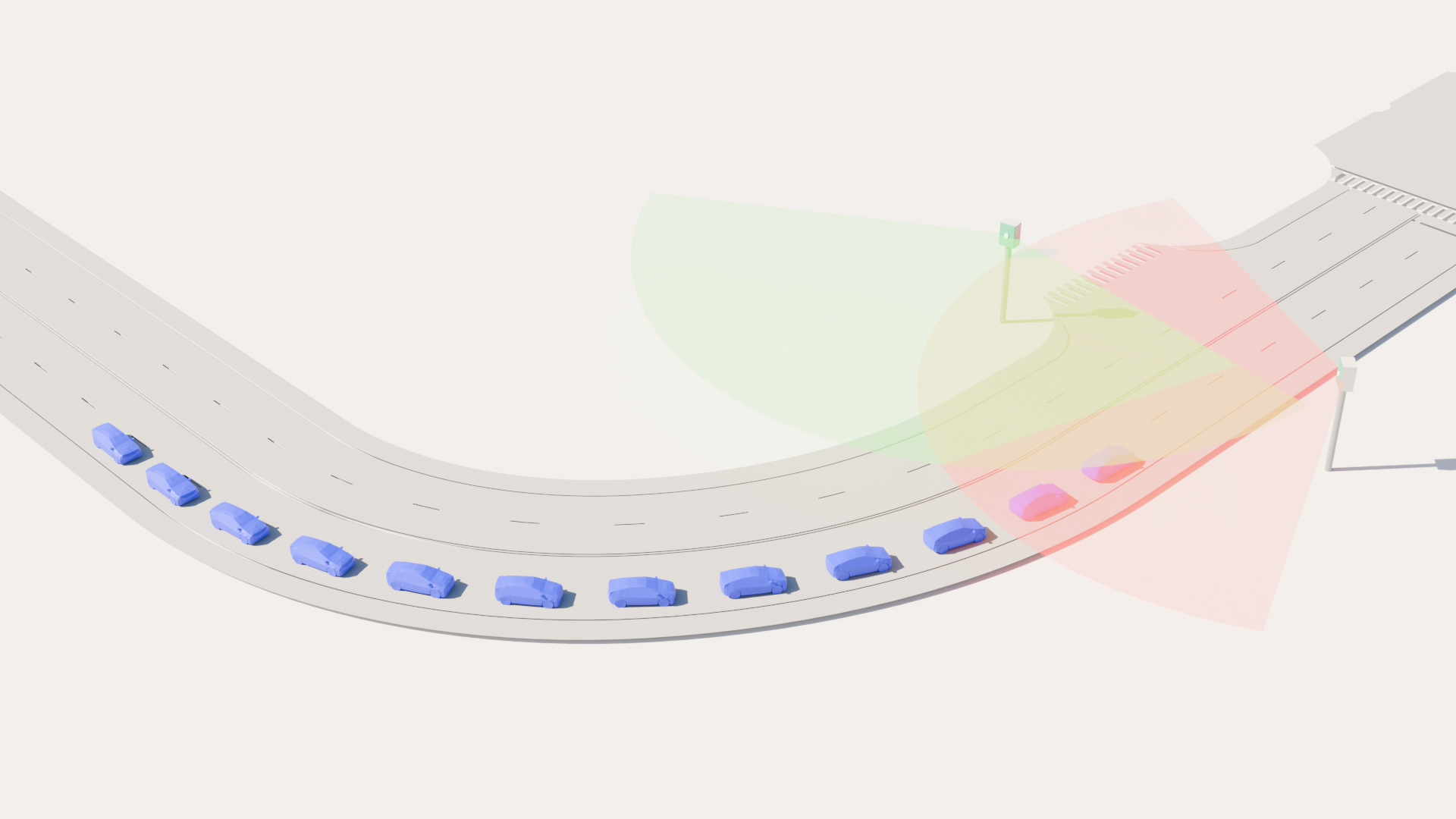}};
        \begin{scope}[x={(image.south east)},y={(image.north west)}]
            \draw[green, thick, ->] (0.6,0.55) node[left=-2pt] {Sensor 1} -- (0.69,0.7);    
            \draw[red, thick, ->] (0.84,0.3) node[left=-2pt] {Sensor 2} -- (0.91,0.53); 
        \end{scope}
    \end{tikzpicture} 
    \includegraphics[width=1\linewidth]{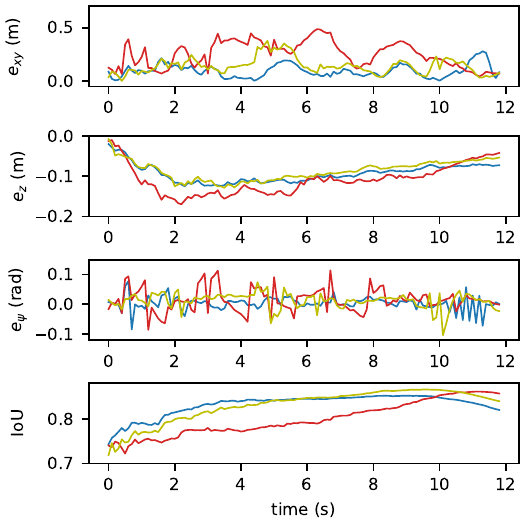}
    \caption{The test scenario in CARLA with a left turn drive and the results. 
    The color blue, red and yellow indicates centralized fusion,  sensor 2, and decentralized fusion, respectively.}
    \label{fig: turn}
\end{figure}
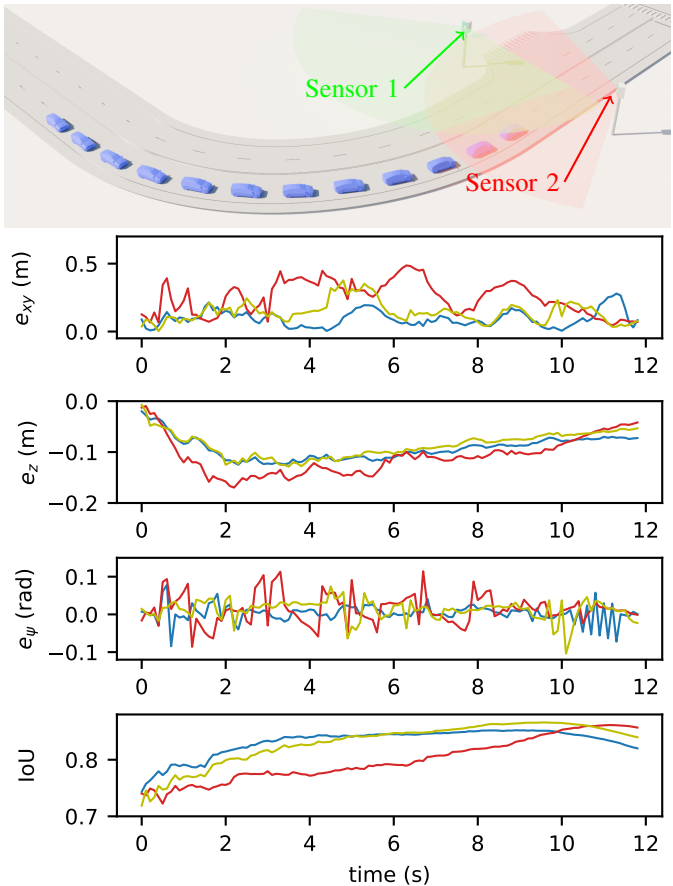

We evaluate the decentralized fusion of the 3D EOT with both simulative data from CARLA \cite{carla} and a realworld open dataset \cite{tumtraf}.
All evaluations use 10 control points for the shape representation and width of $2 \ m$.
The process noise and the measurement noise in the model for all evaluations are set to be the same.
In CARLA, we simulate a vehicle moving along a left turn street segment. 
The drive is observed by 2 lidar sensors, which are placed at the ends of the segment as shown in Fig. \ref{fig: turn}. 
The sensors are installed at the height of 7 $m$ and have the FOV as shown in the figure. 
The sensors operate at 10Hz, generating up to $1e^6$ points per second with 256 layers and the maximum range of 200 $m$. 
The evaluation is conducted with the comparison between the centralized fusion results, where the tracking the conducted using shared data from sensor 1 and sensor 2, the decentralized tracking, where sensor 2 is improved by the sensor 1 given its better perception angle and the tracking with single sensor 2. 


We can see from the figure that the position error in the decentralized fused tracking is significantly reduced compared to the single sensor 2 tracking because it is improved by sensor 1 with its better viewpoint.
However, the centralized tracking is even better and more stable during the whole drive.
The estimation in the vertical direction is similar for all cases, resulting in a maximum error of $-0.2 \ m$.
The error in orientation estimation in the decentralized case is also shown to be reduced compared to the tracking with sensor 2 only.
The Intersection over Union (IoU) measurement is performed in the side view, comparing the estimation with the ground truth mesh. 
We see that the shape estimation gradually improves along the drive, as the measurements become denser.
The centralized fusion performs best at the beginning, as the collected sensor data represent the shape better.
Decentralized tracking outperforms sensor 2 due to the improvement by sensor 1. 
The IoU decreases slightly for the centralized fusion at the end, mainly due to errors in orientation estimation.

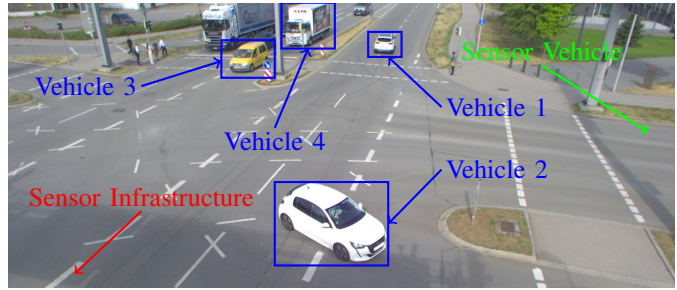
\begin{figure}[!h]
\centering
        \begin{tikzpicture}
        \clip (0,50pt) rectangle (\linewidth,0.625\linewidth);
        \node[anchor=south west,inner sep=0] (image) at (0,0) 
            {\includegraphics[width=\linewidth]{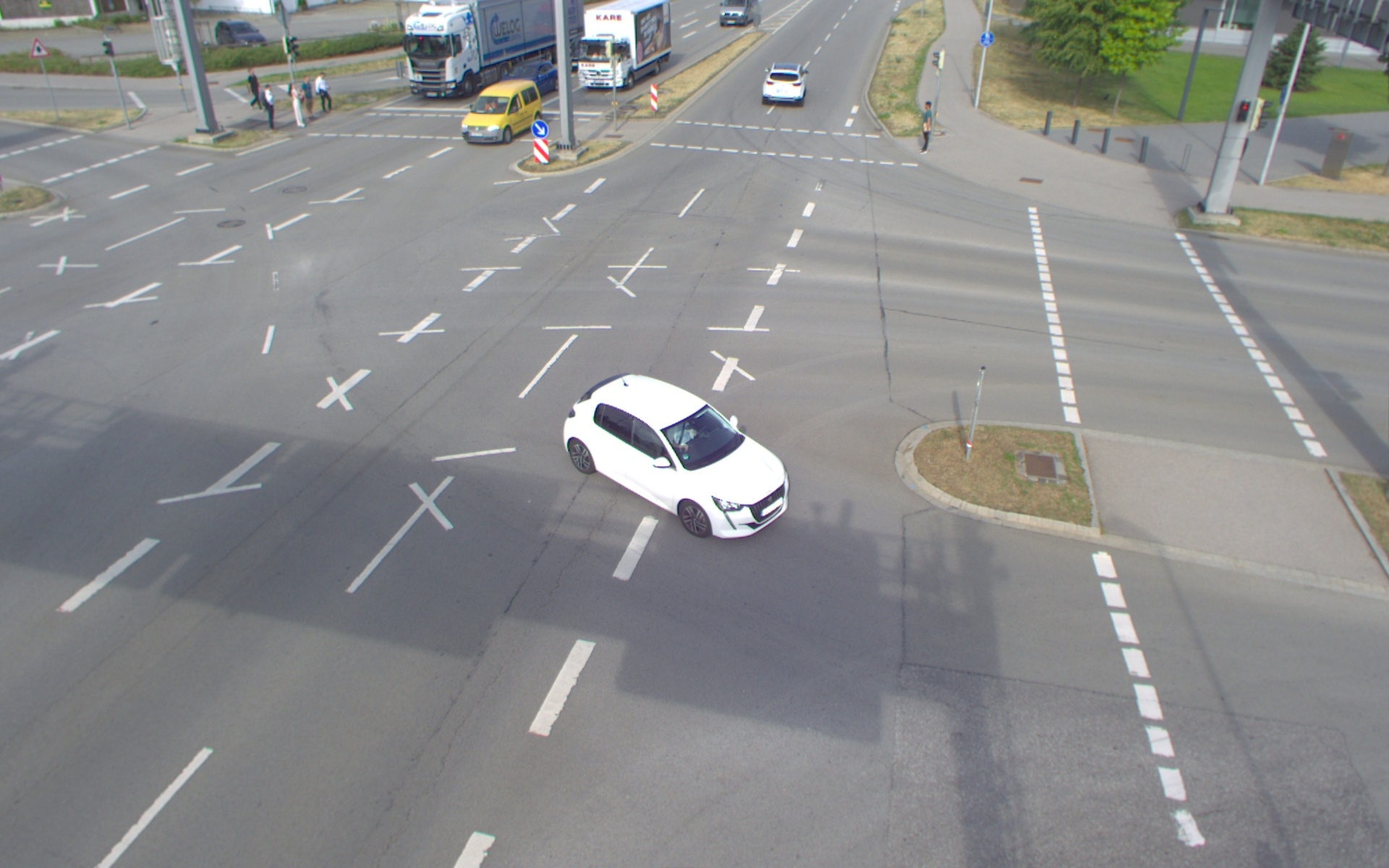}};
        \begin{scope}[x={(image.south east)},y={(image.north west)}]
            \draw[blue, thick, ->] (0.65, 0.75) node[right=-2pt] {Vehicle 1} -- (0.57, 0.87);
            \draw[blue, thick] (0.54,0.87) rectangle (0.59,0.93);
            \draw[blue, thick, ->] (0.65, 0.6) node[right=-2pt] {Vehicle 2} -- (0.57, 0.47);;
            \draw[blue, thick] (0.4,0.37) rectangle (0.57,0.57);
            \draw[blue, thick, ->] (0.2, 0.8) node[left=-2pt] {Vehicle 3} -- (0.32, 0.86);
            \draw[blue, thick] (0.32,0.82) rectangle (0.4,0.91);     
            \draw[blue, thick, ->] (0.4, 0.7) node[below=-2pt] {Vehicle 4} -- (0.45, 0.89);
            \draw[blue, thick] (0.41,0.89) rectangle (0.49,1);
            \draw[red, thick, ->] (0.2, 0.5) node[above=-2pt] {Sensor Infrastructure} -- (0.1, 0.35);
            \draw[green, thick, ->] (0.8, 0.85) node[above=-2pt] {Sensor Vehicle} -- (0.96, 0.69);
            \end{scope}
    \end{tikzpicture} 
    \caption{The test scenario in the real word dataset (sequence R4\_S4) in \cite{tumtraf}.}
    \label{fig:tumtraf vehicles}
\end{figure}

We further evaluate the fusion with the real world dataset \cite{tumtraf}.
In the sequence R4\_S4 of this dataset, 4 turning vehicles at the intersection are observed by both the infrastructure lidar mounted on a gantry and a rooftop lidar on the ego car waiting at the traffic light. 
The scene is shown in Fig. \ref{fig:tumtraf vehicles}.
The infrastructure radar observes the vehicles with less data due to distance and perspective: as the car moves away, only sparse reflections from the rear can be seen.
We use vehicle lidar tracking to improve the infrastructure lidar result.

\begin{figure}[!h]
    \centering
    \includegraphics[trim={0.2cm 0.1cm 0.5cm 1.2cm},clip, width=0.8\linewidth]{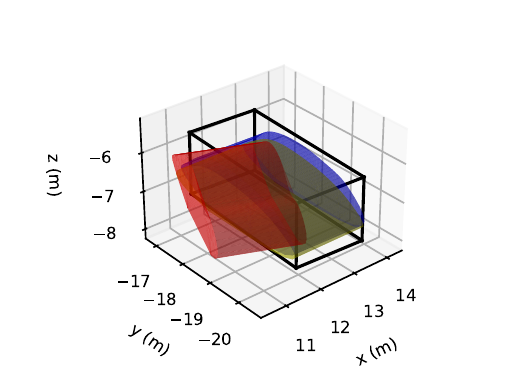}
    \caption{Exampel result of real-world dataset evaluation. 
    Red, yellow, blue results are from infrastructure sensor tracking, decentralized tracking and centralized tracking respectively. 
    The ground truth bounding box is shown in black.}
    \label{fig: example}
\end{figure}

An example of the tracking result is shown in Fig. \ref{fig: example}. 
Here, vehicle 2 is used as a reference. 
The red estimation from the infrastructure sensor provides an oblique result given the side-only observation.
Both centralized and decentralized tracking give good estimates as they fall well within the ground truth bounding box and the shape also qualitatively matches the car.
\begin{figure*}
    \centering
    \includegraphics[width=\linewidth]{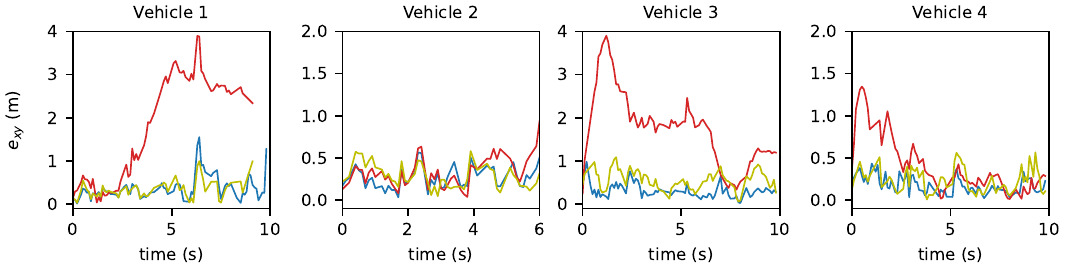}
    \caption{The evaluation result using the TUMTraf dataset.
    The colors blue, red, and yellow indicate centralized tracking, tracking with infrastructure sensor, and decentralized tracking, respectively.
    Vehicle 2 enters the blind zone of the infrastructure sensor after $6 \ s$, so the tracking stops.}
    \label{fig:tumtrafresult}
\end{figure*}
Further result is shown in figure \ref{fig:tumtrafresult}. 
We see that the infrastructure lidar generally suffers from an unfavorable perspective in this scenario.
In all cases, the decentralized fusion can help to improve the result dramatically.
The orientation RMSE for the 4 vehicles are $0.08, 0.06, 0.08, 0.05 \ rad$ for the decentralized tracking, which indicates a reliable orientation estimation.
\section{Conclusion}
\label{chapter: conclusion}
In this work, we investigate decentralized 3D EOT tracking with a shape model that is the extrusion of a side-view profile of the object.
The side view profile is represented by a B-spline curve with flexible control points.
The decentralized fusion is based on CI.
We evaluate the decentralized tracking with simulated data and a real open dataset and show that the fusion significantly helps to improve the tracking result of a sensor that alone has an unfavorable perspective.
Future work includes the analysis of the correlation between the estimates and the investigation of more advanced fusion schemes. 
Furthermore, decentralized fusion will be integrated into hardware environments \cite{flexsense} to study the influence of network connectivity on the results.
\bibliographystyle{IEEEtran}
\bibliography{refseusipco}

\end{document}